\pdfoutput=1

\documentclass[11pt]{article}
\usepackage[]{acl}
\usepackage{times}
\usepackage{latexsym}
\usepackage{booktabs}
\usepackage[T1]{fontenc}
\usepackage[utf8]{inputenc}
\usepackage{microtype}

\usepackage{hyperref}
\usepackage{multicol}
\usepackage{multirow}
\usepackage{makecell}
\usepackage{arydshln}
\newcommand{\ph}[1]{\phantom{#1}}
\makeatletter
\def\adl@drawiv#1#2#3{%
\hskip.5\tabcolsep
\xleaders#3{#2.5\@tempdimb #1{1}#2.5\@tempdimb}%
#2\z@ plus1fil minus1fil\relax
\hskip.5\tabcolsep}
\newcommand{\cdashlinelr}[1]{%
  \noalign{\vskip\aboverulesep
 \global\let\@dashdrawstore\adl@draw
 \global\let\adl@draw\adl@drawiv}
  \cdashline{#1}
  \noalign{\global\let\adl@draw\@dashdrawstore
 \vskip\belowrulesep}}
\makeatother
\usepackage{enumitem}

\title{The Ups and Downs of Large Language Model Inference \\ with Vocabulary Trimming by Language Heuristics}

\author{Nikolay Bogoychev\quad
Pinzhen Chen\quad
Barry Haddow\quad
Alexandra Birch \\
School of Informatics, University of Edinburgh\\
\texttt{\{nbogoych,pinzhen.chen,bhaddow,a.birch\}@ed.ac.uk}
}

\begin{document}
\maketitle
\begin{abstract}
Deploying large language models (LLMs) encounters challenges due to intensive computational and memory requirements. Our research examines vocabulary trimming (VT) inspired by restricting embedding entries to the language of interest to bolster time and memory efficiency. While such modifications have been proven effective in tasks like machine translation, tailoring them to LLMs demands specific modifications given the diverse nature of LLM applications. We apply two language heuristics to trim the full vocabulary---Unicode-based script filtering and corpus-based selection---to different LLM families and sizes. The methods are straightforward, interpretable, and easy to implement. It is found that VT reduces the memory usage of small models by nearly 50\% and has an upper bound of 25\% improvement in generation speed. Yet, we reveal the limitations of these methods in that they do not perform consistently well for each language with diminishing returns in larger models.
\end{abstract}

\section{Introduction}
Large language models (LLMs) are gaining increasing attention given their strong performance \citep{Radford2019Language,brown2020language,workshop2023BLOOM,touvron2023LLaMA}. LLMs, especially multilingual ones, hold vocabulary items for many languages and scripts, which entail a costly matrix multiplication $H \times |V|$ in the output layer, where $H$ is the hidden size and $|V|$ is the size of a vocabulary $V$. This expensive operation leads to increased costs of both memory and time given the autoregressive nature of LLM decoding. Given their substantial size, this latency in inference significantly escalates the expense of LLM deployment.

In practice, creating a sub-vocabulary $V'$ with $|V'|\ll|V|$ and only loading its corresponding embedding entries for inference seems favourable since most logits from the output layer do not affect the hypothesis token(s) at each time step. Vocabulary trimming (VT) has been actively explored in machine translation \citep[often called \textit{shortlisting},][]{schwenk-etal-2007-smooth, le-etal-2012-continuous, devlin-etal-2014-fast}---it computes token-level alignments and makes potential target tokens a sub-vocabulary. While anticipating certain limitations such as domain mismatch \cite{bogoychev-chen-2021-highs,domhan-etal-2022-devil}, vocabulary shortlisting in LLMs poses a fundamental challenge: often LLM outputs are variable and open-ended, complicating the determination of the required lexicons. Recent attempts at multilingual pre-trained models select tokens in a task's language \citep{abdaoui-etal-2020-load,ushio2023efficient}. Nonetheless, research in this direction is still limited, especially in speed considerations.

We follow the idea of fitting vocabulary to the language of the downstream task. Specifically, We examine two strategies: \emph{Unicode-based filtering} where vocabulary items are removed if they do not belong to the task language, and \emph{corpus-based selection} where we record vocabulary hits from a large representative corpus. After experimenting with LLMs from two families of different sizes, we identify a good upper bound of memory reduction with several limitations and outlooks: 1) Unicode-based script filtering maintains quality for Latin-based languages but harms languages requiring code-mixing. 2) Corpus-based selection leads to fewer alterations but is less effective in reducing the embedding size. 3) Embeddings are proportionally smaller in larger models (with smaller vocabularies). Yet we argue that VT can be applied orthogonally to other efficiency methods like efficient attention, quantization, etc.

\section{Language-Based Vocabulary Trimming}
We explore two ways to prepare sub-vocabulary for LLMs, focusing on only retaining tokens relevant to the language being generated. We test a batched setting on the fly: we determine a sub-vocabulary for an entire batch because creating the sub-vocabulary separately for each input is too expensive in practice. Furthermore, we always include all tokens appearing in the inputs.

\paragraph{Script-based filtering}
This is done by filtering token strings that fall out of a language's Unicode range---keeping tokens in the writing script of that language. It should be especially effective for languages operating on unique scripts, such as Armenian, Chinese, Korean, etc since it allows for concise vocabulary restriction. This method might be less practical if a writing system is shared among multiple languages (e.g. Cyrillic or Latin alphabets), because it would be infeasible to limit the lexicons to those solely belong to a specific language, resulting in a relatively large sub-vocabulary. Moreover, this method would strictly rule out code-mixed texts, emojis, etc, which are used in real-world communications.

\paragraph{Corpus-based selection}
Another way is to tokenize a representative corpus in the desired language in advance and use the vocabulary entries that have been recorded to build a sub-vocabulary. This method is non-exhaustive because we could miss rare but valid tokens or suffer from domain mismatch between the vocabulary selection corpus and the downstream tasks at inference time.

\section{Experimental Setup}

\paragraph{Languages and test sets}
We experimented on four languages: Bulgarian, Chinese, English, and Spanish, to offer distinct conditions that cover different degrees of writing script overlap, code-mixing, etc, with details in Appendix~\ref{sec:languages}. We sample 50 prompt questions from OpenAssistant \citep{kopf2304openassistant} which are then human-translated into test languages. We decode them with beam size 1.

\paragraph{Metrics}
We consider efficiency-quality trade-offs. In terms of speed, we report end-to-end time to decode the entire test, including model loading and embedding slicing. As a quality indicator, we count the chances a model fails to produce \textit{the exact same output} (miss) with a full vocabulary and with VT. In addition, we report the BLEU and chrF of the VT output w.r.t. to the \emph{original output} with the full vocabulary (not the reference). Note that there is no gold reference due to the open domain nature; we hence prefix the two string metrics with an ``o-''.

\paragraph{Large language models}
We experiment with instruction-tuned LLMs based on BLOOM at various sizes \citep{workshop2023BLOOM} as well as LLaMA-7B \citep{touvron2023LLaMA}. We adopt \citet{chen2023monolingual}'s models fine-tuned on machine translations of the Alpaca dataset \citep{alpaca} to test for open domain question answering, which maximizes the difficulty for VT as explained earlier.

BLOOM is multilingual and explicitly supports English, Spanish, and Chinese, but not Bulgarian. Consequently, it has a sizeable vocabulary of 250K and is therefore a prime and tempting candidate to reduce vocabulary for a specific language during inference. We experiment with the 560M, 1B7, and 7B1 checkpoints, with diminishing computational burden on the embedding and output layers.

LLaMA is an English-centric LLM with a small 32K vocabulary. We might have reduced benefit from VT because a proportionally lower amount of computation occurs in the output layer. On the other hand, since the LLM is European language-focused, we expect drastic vocabulary reductions compared to BLOOM for Bulgarian and Chinese.

\paragraph{Vocabulary trimming details}
We tokenize the test prompts and always include input tokens in the sub-vocabulary. We then apply either of the proposed selection methods. Script-based filtering checks whether a vocabulary entry belongs to a Unicode subset: Cyrillic for Bulgarian, ASCII for English, Latin Extended-A for Spanish, and Chinese characters for Chinese. Whereas for corpus-based selection, we tokenize a subset of WikiMatrix \citep{schwenk-etal-2021-wikimatrix} containing Wikipedia texts for each language and record vocabulary hits.

For both selection methods, we keep the first 300 vocabulary entries of each LLM too, as those usually correspond to special tokens, Unicode bytes (for byte-level BPE), numbers, etc. We compute the sub-vocabulary offline and we do not record the time spent on pre-tokenizing a large corpus or extracting a Unicode subset in the measurements, as once done, these can be reused for every batch during inference. Script-based filtering takes under 60 seconds and corpus-based selection takes up to 10 minutes. Adding the inputs' tokens to the sub-vocabulary takes negligible time.

\begin{table*}[ht]
\centering\small
\setlength{\tabcolsep}{0.8ex}
\begin{tabular}{lrcccccccccccc}
\toprule
\multicolumn{1}{c}{\multirow[b]{2}{*}{Language}} &  \multicolumn{1}{c}{\multirow[b]{2}{*}{$|V|$}} & \multicolumn{4}{c}{{BLOOM-560M}} & \multicolumn{4}{c}{{BLOOM-1B7}} & \multicolumn{4}{c}{{BLOOM-7B1}} \\ 
\cmidrule(lr){3-6}\cmidrule(lr){7-10}\cmidrule(lr){11-14}
 & & time & miss & o-BLEU & o-chrF & time & miss & o-BLEU & o-chrF & time & miss & o-BLEU & o-chrF \\
\midrule
\textbf{bg} full & 250680\hspace{1ex} & 05:26 & -- & & & 15:18 & -- & & & 65:01 & -- & & \\
\hspace{1ex}Unicode & 22912\hspace{1ex} & 04:39 & 1 & 99.04 & 99.73 & 13:44 & 4 & 91.67 & 96.36 & 51:46 & 10 & 83.42 & 87.03 \\
\hspace{1ex}corpus  & 58642\hspace{1ex} & 04:49 & 0 &       &       & 09:34 & 1 & 99.49 & 99.85 & 60:28 & 3  & 91.68 & 95.84 \\
\hspace{1ex}\textit{oracle} & 1408\hspace{1ex} & 04:22 & 0  & &  & 12:31  & 0 & & & 61:06 & 0 & &  \\
\midrule
\textbf{en} full & 250680\hspace{1ex} & 07:37 & --   & & & 16:35 & -- & &  & 55:08 & -- & & \\
\hspace{1ex}Unicode & 186752\hspace{1ex} & 07:40 & 1 & 98.21 & 98.68 & 16:05 & 0 & & & 58:18 & 0  & & \\
\hspace{1ex}corpus & 113024\hspace{1ex} & 07:00 & 1  & 99.22 & 99.45 & 15:08 & 3 & 96.59 & 98.88 & 54:20 & 2 & 98.91 & 99.40 \\
\hspace{1ex}\textit{oracle} & 4736\hspace{1ex} & 06:14 & 0  & & & 13:06 & 0 & & &  48:46 & 0 & &  \\
\midrule
\textbf{es} full & 250680\hspace{1ex} & 05:58   & -- &  &  & 12:26 & -- & &  & 63.15 & -- & & \\
\hspace{1ex}Unicode & 187008\hspace{1ex} & 05:48 & 0 & & & 12:01 & 0 & & & 59:15 & 0 & &  \\
\hspace{1ex}corpus  & 112128\hspace{1ex} & 05:37 & 0 & &  & 11:34 & 4 & 95.91 & 97.71 & 57:41 & 4 & 94.46 & 96.25 \\
\hspace{1ex}\textit{oracle}  & 4736\hspace{1ex} & 04:53 & 0 & &   & 09:26 & 0 & & & 51:43 & 0 & & \\
\midrule
\textbf{zh} full & 250680\hspace{1ex} & 06:29 & -- & &  & 15:27 & -- & &  & 55:09 & -- & & \\
\hspace{1ex}Unicode & 51584\hspace{1ex}  & 05:54 & 16 & 53.32 & 70.38 & 13:09 & 21 & 47.66 & 63.66 & 50:50 & 22 & 47.76 & 63.60 \\
\hspace{1ex}corpus  & 104320\hspace{1ex} & 06:08 & 11 & 66.17 & 78.44 & 14:08 & 16 & 63.26 & 73.99 & 46:39 & 17 & 62.37 & 76.55 \\
\hspace{1ex}\textit{oracle} & 4096\hspace{1ex} & 05:16 & 0 & & & 12:07 & 0 & & & 50:50  & 0  & & \\
\bottomrule
\end{tabular}
\caption{CPU VT results for the BLOOM family.}
\label{tab:results_BLOOM}
\end{table*}

\begin{table}[ht]
\centering\small
\setlength{\tabcolsep}{1ex}
\begin{tabular}{lrcccc}
\toprule
\multicolumn{1}{c}{\multirow[b]{2}{*}{Language}} &  \multicolumn{1}{c}{\multirow[b]{2}{*}{$|V|$}} & \multicolumn{4}{c}{LLaMA-7B} \\ 
\cmidrule(lr){3-6}
 & & time & miss & o-BLEU & o-chrF \\
\midrule
\textbf{bg} full & 32000 & 117:15 & -- \\
\hspace{1ex}Unicode & 4736  & 125:55 & 19 & 74.13 & 81.51 \\
\hspace{1ex}corpus  & 26496 & 132:24 & 5 & 97.75 & 98.44 \\
\hspace{1ex}\textit{oracle}  & 2048  & 123:06 & 0 \\
\midrule
\textbf{en} full & 32000 & 113:52 & -- \\
\hspace{1ex}Unicode & 27520 & 125:57 & 6 & 79.43 & 88.91 \\
\hspace{1ex}corpus  & 30720 & 111:30 & 19 & 93.07 & 97.14 \\
\hspace{1ex}\textit{oracle}  & 4480  & 119:32 & 0  \\
\midrule
\textbf{es} full & 32000 & 131:03 & -- \\
\hspace{1ex}Unicode & 27648 & 128:00 & 8 & 89.60 & 91.62 \\
\hspace{1ex}corpus  & 30336 & 129:26 & 2 & 97.08 & 99.15 \\ 
\hspace{1ex}\textit{oracle}  & 3456  & 123:25 & 0  \\
\midrule
\textbf{zh} full & 32000 & 130:42  & -- \\
\hspace{1ex}Unicode & 2688  & 114:39  & 13 & 75.43 & 83.36 \\
\hspace{1ex}corpus  & 28160 & 119:58  & 2 & 95.47 & 97.80 \\
\hspace{1ex}\textit{oracle}  & 1536  & 126:16  & 0 \\
\bottomrule
\end{tabular}
\caption{CPU VT results for LLaMA-7B.}
\label{tab:results_LLaMA}
\end{table}

\paragraph{Hardware}
We conduct experiments both on CPU and GPU devices. For the CPU tests, we use Xeon Gold 6248 (40 Cores, 80 Threads), and for GPU tests, we use a single Nvidia RTX 3090. CPU inference is performed in float32 precision, whereas GPU inference is in int8 \citep{dettmers2022llmint8}.

\section{CPU Results and Discussions}

\paragraph{Upper bound performance} First of all, we conduct an \emph{oracle} vocabulary selection experiment to find the theoretical upper bound for speed and memory improvements: we run inference using the full vocabulary and we add the used vocabulary items to a \emph{oracle} sub-vocabulary.

\paragraph{BLOOM versus LLaMA} We present CPU results for the BLOOM family in Table~\ref{tab:results_BLOOM} and those for LLaMA-7B in Table~\ref{tab:results_LLaMA}. We observe around 20\% time improvements with the smaller BLOOM at 560M and 1B7, but only 5--10\% in the 7B models. As the model grows in size, the oracle upper bound sees decreasing gains, due to the proportion of the embedding matrices becoming smaller in a larger model. By comparing BLOOM-7B1 with LLaMA-7B, we also find that the larger the base vocabulary, the more effective VT is. The oracle vocabulary is more than an order of magnitude smaller than our VT approaches, but in practice, it would be difficult to reduce the vocabulary size by as much.

Speed numbers of LLaMA-7B on CPU are relatively inconsistent and had wide variance across test runs. We attribute this to the small vocabulary size and thus less computational footprint in the output layer affected by VT. Also, there could be various scheduling issues and non-deterministic cache accesses as GEMM operations are split across the 40 cores of the CPU.

\subsection{Script-based vocabulary trimming}

When applying script-based filtering, we observe different trends in English and Spanish compared to Bulgarian and Chinese. For BLOOM, the sub-vocabulary size for Bulgarian and Chinese can be reduced to 10--20\%, whereas for English and Spanish, it remains at 60\%. This is potentially because BLOOM allocated more vocabulary items for European languages which are the dominant ones when the tokenizer is trained. Generally, the inference time reduces to between full and oracle vocabulary. In terms of misses, the model can maintain almost the same outputs with and without VT for English and Spanish. However, there are 10--20\% misses for Bulgarian and 30--40\% for Chinese. 

LLaMA-7B results are less favourable: script-based filtering does not significantly reduce the vocabulary size for English and Spanish, and all languages suffer from relatively high misses between 10--40\%. Specifically for Bulgarian and Chinese, we argue that Unicode filtering could be too harsh as sometimes English characters are code-mixed in the language and cannot be avoided, e.g., when generating a website link. Therefore, we conclude that VT based on the writing script can improve inference efficiency without degrading performance for a multilingual LLM to generate Latin languages, but it is less feasible for non-Latin languages or English-centric LLMs with a smaller vocabulary.

\subsection{Corpus-based vocabulary trimming}
Corpus-based selection leaves a much larger vocabulary for Bulgarian and Chinese but reduces the vocabulary to half or less for English and Spanish. This method produces a more balanced sub-vocabulary for each language likely due to the inclusion of tokens outside of the desired language. However, for LLaMA-7B which has a small vocabulary in the first place, this approach keeps most of the entries for all languages and is thus not useful.

The corpus-based selection also ameliorates the quality problem to some extent by allowing for code-mixing (usually English), although the Chinese VT models still struggle to produce identical output as the full vocabulary models. Overall, we see a small but consistent reduction in runtime with BLOOM for this VT approach, indicating its practicality at least for English.

\subsection{Memory}
Besides speed considerations, VT can lead to ample memory footprint reduction, especially for smaller models like BLOOM-560M, where the model size is dominated by the vocabulary (nearly 50\% of all model parameters). In practice, these models are small enough to fit in modern GPUs and CPUs, so the reduced memory is not game-changing. On the other hand, when looking at bigger models like BLOOM-7B1 or LLaMA-7B, vocabulary makes up just a tiny portion of the overall number of parameters and thus the relative reduction in model size is modest and could not enable the use of smaller GPUs. We can view this as a proxy judgement about the computational distribution of the model: The larger the model, the less time is spent in the output layer, and thus the smaller the impact of VT is. Exact memory numbers are available in Table~\ref{tab:memory}.

\begin{table}[thb]
\centering\small
\setlength{\tabcolsep}{0.55ex}
\begin{tabular}{lrcccrc}
\toprule
\multicolumn{1}{c}{\multirow[b]{2}{*}{Language}} & \multicolumn{4}{c}{BLOOM} & \multicolumn{2}{c}{LLaMA} \\
\cmidrule(lr){2-5}\cmidrule(lr){6-7}
 & \multicolumn{1}{c}{$|V|$} & 560M & 1B7 & 7B1 & \multicolumn{1}{c}{$|V|$} & 7B \\
\midrule
Full model & \multicolumn{1}{c}{250680}  & 2.10  & 6.10  & 27.10\ph{0} & 32000 & 27.10\ph{0} \\
\midrule
\multicolumn{7}{l}{Embedding matrix or output layer}\\
\hspace{1ex}full vocab & 250680 & 0.90 & 1.90 & 3.80  & \ph{0}32000 & 0.50 \\
\cdashlinelr{1-7}
\hspace{1ex}bg Unicode  & 22912& 0.09 & 0.18  & 0.36 & 4736 & 0.07 \\
\hspace{1ex}bg corpus & 58642 & 0.22 & 0.45  & 0.90  & 26496 & 0.41 \\
\hspace{1ex}en Unicode & 186752 & 0.70 & 1.40 & 2.80 & 27520 & 0.43 \\
\hspace{1ex}en corpus & 113024 & 0.44 & 0.88  & 1.70 & 30720 & 0.48 \\
\hspace{1ex}es Unicode & 187008 & 0.70 & 1.40 & 2.80 & 27648 & 0.43 \\
\hspace{1ex}es corpus & 112128 & 0.43 & 0.86  & 1.70 & 30336 & 0.47 \\
\hspace{1ex}zh Unicode & 51584 & 0.20 & 0.40 & 0.80 & 2688 & 0.04 \\
\hspace{1ex}zh corpus & 104320 & 0.40 & 0.80 & 1.60 & 28160 & 0.44 \\
\bottomrule
\end{tabular}
\caption{Theoretical memory footprint (in GB) for BLOOM and LLaMA with float32 featuring the embedding matrix.}
\label{tab:memory}
\end{table}

\section{GPU results}
In addition to CPU tests, we performed the same BLOOM experiments on a GPU and observed that all three selection criteria including the oracle do not lead to improved inference speed. Small performance differences might amount to little more than noise when the overhead of model slicing is considered. We hypothesize that GPUs are designed for multiplying large matrices, so reducing the matrix size, even to the extremity of an oracle sub-vocabulary, is not able to offer any speedup. This is consistent with \citet{bogoychev-etal-2020-edinburghs}'s findings in applying shortlists to neural machine translation on GPUs. We list GPU results for BLOOM in Appendix~\ref{sec:gpu-performance} Table~\ref{tab:results_BLOOM_gpu}.

\section{Conclusion}
We presented a study of two straightforward language-inspired vocabulary trimming methods to speed up inference and save memory for large language model deployment. Experiments reveal ups and downs. While we can achieve speed improvements, it does not guarantee that the output is not altered compared to full vocabulary generation. With the models tested, we see the feasibility of our proposed approaches for English and Spanish, but there are shortcomings when considering languages written in non-Latin script and requiring code-mixing. In terms of efficiency, the reduction in inference time is less pronounced compared with memory saving.

\section*{Ethical Considerations}
Our study aimed solely at reducing the computational resource consumption for deploying large language models. Our analysis contributes to the understanding of language heuristics in trimming an LLM vocabulary. While there is minimal risk associated with generating harmful content, it is no different for other research on large language models. We believe research into this direction has a positive impact in terms of energy saving and service deployment.

\section*{Acknowledgement}
This work has received funding from UK Research and Innovation (UKRI) under the UK government’s Horizon Europe funding guarantee [grant numbers 10052546 and 10039436].

\bibliography{custom}

\begin{thebibliography}{17}
\expandafter\ifx\csname natexlab\endcsname\relax\def\natexlab#1{#1}\fi

\bibitem[{Abdaoui et~al.(2020)Abdaoui, Pradel, and Sigel}]{abdaoui-etal-2020-load}
Amine Abdaoui, Camille Pradel, and Gr{\'e}goire Sigel. 2020.
\newblock \href {https://doi.org/10.18653/v1/2020.sustainlp-1.16} {Load what you need: Smaller versions of mutililingual {BERT}}.
\newblock In \emph{Proceedings of SustaiNLP: Workshop on Simple and Efficient Natural Language Processing}.

\bibitem[{Bogoychev and Chen(2021)}]{bogoychev-chen-2021-highs}
Nikolay Bogoychev and Pinzhen Chen. 2021.
\newblock \href {https://doi.org/10.18653/v1/2021.insights-1.12} {The highs and lows of simple lexical domain adaptation approaches for neural machine translation}.
\newblock In \emph{Proceedings of the Second Workshop on Insights from Negative Results in NLP}.

\bibitem[{Bogoychev et~al.(2020)Bogoychev, Grundkiewicz, Aji, Behnke, Heafield, Kashyap, Farsarakis, and Chudyk}]{bogoychev-etal-2020-edinburghs}
Nikolay Bogoychev, Roman Grundkiewicz, Alham~Fikri Aji, Maximiliana Behnke, Kenneth Heafield, Sidharth Kashyap, Emmanouil-Ioannis Farsarakis, and Mateusz Chudyk. 2020.
\newblock \href {https://www.aclweb.org/anthology/2020.ngt-1.26} {{E}dinburgh{'}s submissions to the 2020 machine translation efficiency task}.
\newblock In \emph{Proceedings of the Fourth Workshop on Neural Generation and Translation}.

\bibitem[{Brown et~al.(2020)Brown, Mann, Ryder, Subbiah, Kaplan, Dhariwal, Neelakantan, Shyam, Sastry, Askell et~al.}]{brown2020language}
Tom Brown, Benjamin Mann, Nick Ryder, Melanie Subbiah, Jared~D Kaplan, Prafulla Dhariwal, Arvind Neelakantan, Pranav Shyam, Girish Sastry, Amanda Askell, et~al. 2020.
\newblock \href {https://proceedings.neurips.cc/paper_files/paper/2020/file/1457c0d6bfcb4967418bfb8ac142f64a-Paper.pdf} {Language models are few-shot learners}.
\newblock In \emph{Advances in Neural Information Processing Systems}.

\bibitem[{Chen et~al.(2024)Chen, Ji, Bogoychev, Kutuzov, Haddow, and Heafield}]{chen2023monolingual}
Pinzhen Chen, Shaoxiong Ji, Nikolay Bogoychev, Andrey Kutuzov, Barry Haddow, and Kenneth Heafield. 2024.
\newblock \href {https://aclanthology.org/2024.findings-eacl.90} {Monolingual or multilingual instruction tuning: Which makes a better alpaca}.
\newblock In \emph{Findings of the Association for Computational Linguistics: EACL 2024}.

\bibitem[{Dettmers et~al.(2022)Dettmers, Lewis, Belkada, and Zettlemoyer}]{dettmers2022llmint8}
Tim Dettmers, Mike Lewis, Younes Belkada, and Luke Zettlemoyer. 2022.
\newblock Llm.int8(): 8-bit matrix multiplication for transformers at scale.
\newblock \emph{arXiv preprint arXiv:2208.07339}.

\bibitem[{Devlin et~al.(2014)Devlin, Zbib, Huang, Lamar, Schwartz, and Makhoul}]{devlin-etal-2014-fast}
Jacob Devlin, Rabih Zbib, Zhongqiang Huang, Thomas Lamar, Richard Schwartz, and John Makhoul. 2014.
\newblock \href {https://doi.org/10.3115/v1/P14-1129} {Fast and robust neural network joint models for statistical machine translation}.
\newblock In \emph{Proceedings of the 52nd Annual Meeting of the Association for Computational Linguistics (Volume 1: Long Papers)}.

\bibitem[{Domhan et~al.(2022)Domhan, Hasler, Tran, Trenous, Byrne, and Hieber}]{domhan-etal-2022-devil}
Tobias Domhan, Eva Hasler, Ke~Tran, Sony Trenous, Bill Byrne, and Felix Hieber. 2022.
\newblock \href {https://doi.org/10.18653/v1/2022.naacl-main.136} {The devil is in the details: On the pitfalls of vocabulary selection in neural machine translation}.
\newblock In \emph{Proceedings of the 2022 Conference of the North American Chapter of the Association for Computational Linguistics: Human Language Technologies}.

\bibitem[{K{\"o}pf et~al.(2023)K{\"o}pf, Kilcher, von R{\"u}tte, Anagnostidis, Tam, Stevens, Barhoum, Duc, Stanley, Nagyfi et~al.}]{kopf2304openassistant}
Andreas K{\"o}pf, Yannic Kilcher, Dimitri von R{\"u}tte, Sotiris Anagnostidis, Zhi-Rui Tam, Keith Stevens, Abdullah Barhoum, Nguyen~Minh Duc, Oliver Stanley, Rich{\'a}rd Nagyfi, et~al. 2023.
\newblock \href {https://openreview.net/forum?id=VSJotgbPHF} {Openassistant conversations - democratizing large language model alignment}.
\newblock In \emph{Thirty-seventh Conference on Neural Information Processing Systems Datasets and Benchmarks Track}.

\bibitem[{Le et~al.(2012)Le, Allauzen, and Yvon}]{le-etal-2012-continuous}
Hai~Son Le, Alexandre Allauzen, and Fran{\c{c}}ois Yvon. 2012.
\newblock \href {https://aclanthology.org/N12-1005} {Continuous space translation models with neural networks}.
\newblock In \emph{Proceedings of the 2012 Conference of the North {A}merican Chapter of the Association for Computational Linguistics: Human Language Technologies}.

\bibitem[{Radford et~al.(2019)Radford, Wu, Child, Luan, Amodei, and Sutskever}]{Radford2019Language}
Alec Radford, Jeff Wu, Rewon Child, David Luan, Dario Amodei, and Ilya Sutskever. 2019.
\newblock \href {https://cdn.openai.com/better-language-models/language_models_are_unsupervised_multitask_learners.pdf} {Language models are unsupervised multitask learners}.
\newblock Openai.com.

\bibitem[{Scao et~al.(2022)Scao, Fan, Akiki, Pavlick, Ili{\'c}, Hesslow, Castagn{\'e}, Luccioni, Yvon, Gall{\'e} et~al.}]{workshop2023BLOOM}
Teven~Le Scao, Angela Fan, Christopher Akiki, Ellie Pavlick, Suzana Ili{\'c}, Daniel Hesslow, Roman Castagn{\'e}, Alexandra~Sasha Luccioni, Fran{\c{c}}ois Yvon, Matthias Gall{\'e}, et~al. 2022.
\newblock \href {https://arxiv.org/abs/2211.05100} {Bloom: A {176B}-parameter open-access multilingual language model}.
\newblock \emph{arXiv preprint arXiv:2211.05100}.

\bibitem[{Schwenk et~al.(2021)Schwenk, Chaudhary, Sun, Gong, and Guzm{\'a}n}]{schwenk-etal-2021-wikimatrix}
Holger Schwenk, Vishrav Chaudhary, Shuo Sun, Hongyu Gong, and Francisco Guzm{\'a}n. 2021.
\newblock \href {https://doi.org/10.18653/v1/2021.eacl-main.115} {{W}iki{M}atrix: Mining 135{M} parallel sentences in 1620 language pairs from {W}ikipedia}.
\newblock In \emph{Proceedings of the 16th Conference of the European Chapter of the Association for Computational Linguistics: Main Volume}.

\bibitem[{Schwenk et~al.(2007)Schwenk, R.~Costa-juss{\`a}, and R.~Fonollosa}]{schwenk-etal-2007-smooth}
Holger Schwenk, Marta R.~Costa-juss{\`a}, and Jose~A. R.~Fonollosa. 2007.
\newblock \href {https://aclanthology.org/D07-1045} {Smooth bilingual $n$-gram translation}.
\newblock In \emph{Proceedings of the 2007 Joint Conference on Empirical Methods in Natural Language Processing and Computational Natural Language Learning ({EMNLP}-{C}o{NLL})}.

\bibitem[{Taori et~al.(2023)Taori, Gulrajani, Zhang, Dubois, Li, Guestrin, Liang, and Hashimoto}]{alpaca}
Rohan Taori, Ishaan Gulrajani, Tianyi Zhang, Yann Dubois, Xuechen Li, Carlos Guestrin, Percy Liang, and Tatsunori~B. Hashimoto. 2023.
\newblock \href {https://github.com/tatsu-lab/stanford_alpaca} {Stanford alpaca: An instruction-following llama model}.
\newblock GitHub repository.

\bibitem[{Touvron et~al.(2023)Touvron, Lavril, Izacard, Martinet, Lachaux, Lacroix, Rozi{\`e}re, Goyal, Hambro, Azhar et~al.}]{touvron2023LLaMA}
Hugo Touvron, Thibaut Lavril, Gautier Izacard, Xavier Martinet, Marie-Anne Lachaux, Timoth{\'e}e Lacroix, Baptiste Rozi{\`e}re, Naman Goyal, Eric Hambro, Faisal Azhar, et~al. 2023.
\newblock \href {https://arxiv.org/abs/2302.13971} {{LLaMA}: Open and efficient foundation language models}.
\newblock \emph{arXiv preprint arXiv:2302.13971}.

\bibitem[{Ushio et~al.(2023)Ushio, Zhou, and Camacho-Collados}]{ushio2023efficient}
Asahi Ushio, Yi~Zhou, and Jose Camacho-Collados. 2023.
\newblock \href {https://doi.org/10.18653/v1/2023.findings-emnlp.981} {Efficient multilingual language model compression through vocabulary trimming}.
\newblock In \emph{Findings of the Association for Computational Linguistics: EMNLP 2023}.

\end{thebibliography}
\bibliographystyle{acl}

\appendix

\section{Languages}\label{sec:languages}
We experimented with Bulgarian, Chinese, English, and Spanish, to cover different conditions and use cases regarding writing scripts and text usage. English and Spanish use the same script and have a high overlap in vocabulary with many other languages after tokenization. Since LLMs are English-centric, we examine how effective of a sub-vocabulary we can find when it is not possible to shortlist merely based on the script. Bulgarian is a low-resource language written in the Cyrillic script. Most multilingual language models have lower amounts of Cyrillic tokens, so we expect that script-based filtering will leave a small sub-vocabulary; however, since Cyrillic is used by other languages, we will inevitably end up with vocabulary items that do not belong to Bulgarian. Finally, Chinese is a high-resource language with a unique script; Unicode filtering would be the most effective in this case.

\section{GPU Performance}\label{sec:gpu-performance}
We provide GPU performance numbers in Table~\ref{tab:results_BLOOM_gpu}. Unfortunately, neither of the language-based vocabulary trimming methods can improve time efficiency.

\begin{table}[hb]
\centering\small
\setlength{\tabcolsep}{0.55ex}
\begin{tabular}{lrcccccc}
\toprule
\multicolumn{1}{c}{\multirow[b]{2}{*}{Language}} &  \multicolumn{1}{c}{\multirow[b]{2}{*}{$|V|$}} & \multicolumn{2}{c}{\makecell{BLOOM\\-560M}} & \multicolumn{2}{c}{\makecell{BLOOM\\-1B7}} & \multicolumn{2}{c}{\makecell{BLOOM\\-7B1}} \\ 
\cmidrule(lr){3-4}\cmidrule(lr){5-6}\cmidrule(lr){7-8}
 & & time & miss & time & miss & time & miss \\
\midrule
\textbf{bg} full & 250680\hspace{1ex} & 05:22 & -- & 08:29 & -- & 14:43 & --\\
\hspace{1ex}Unicode & 22912\hspace{1ex}  & 05:23 & 0 & 08:45 & 6& 14:35 & 17  \\
\hspace{1ex}corpus  & 58642\hspace{1ex}  & 05:22 & 0 &  08:38 &  1  &  14:33 &10 \\
\hspace{1ex}\textit{oracle} & 1408\hspace{1ex} & 05:21 & 0  & 09:06  & 0& 14:33 & 0 \\
\midrule
\textbf{en} full & 250680\hspace{1ex} & 06:50 & -- & 09:02 & -- & 11:54 & --\\
\hspace{1ex}Unicode & 186752\hspace{1ex} & 06:54 & 0 & 08:52 & 0& 11:46 & 0 \\
\hspace{1ex}corpus  & 113024\hspace{1ex} & 06:38 & 2 & 08:56 & 3& 11:59 & 3 \\
\hspace{1ex}\textit{oracle} & 4736\hspace{1ex} & 06:43 & 0 & 09:00 & 0&  11:52 & 0 \\
\midrule
\textbf{es} full & 250680\hspace{1ex} & 06:17 & -- & 07:05 & -- & 12:35 & --\\
\hspace{1ex}Unicode & 187008\hspace{1ex} & 06:13 & 0 & 07:03 & 0& 12:17 & 0 \\
\hspace{1ex}corpus  & 112128\hspace{1ex} & 06:15 & 1 & 7:10 & 3& 12:30 & 3 \\
\hspace{1ex}\textit{oracle}  & 4736\hspace{1ex} & 06:26 & 0  & 07:23 & 0& 12:17 & 0
\\
\midrule
\textbf{zh} full & 250680\hspace{1ex} & 05:37 & -- & 08:47 & -- & 11:58 & --\\
\hspace{1ex}Unicode & 51584\hspace{1ex}  & 06:10 & 15  & 08:34 & 20 & 11:22 & 29  \\
\hspace{1ex}corpus  & 104320\hspace{1ex} & 06:03 & 11  & 09:01 & 16 & 11:35 & 13  \\
\hspace{1ex}\textit{oracle} & 4096\hspace{1ex} & 05:35 & 0 & 08:42 & 0& 11:46  & 0 \\
\bottomrule
\end{tabular}
\caption{GPU VT results for the BLOOM family.}
\label{tab:results_BLOOM_gpu}
\end{table}

\end{document}